\documentclass[letterpaper, 10 pt, conference]{ieeeconf}  
\usepackage{graphicx}
\usepackage{amsmath}
\usepackage{amssymb}
\usepackage{amsfonts}
\usepackage{bm}
\usepackage{booktabs} 
\usepackage{multirow}
\usepackage{xcolor}
\usepackage{float}
\providecommand{\IEEEoverridecommandlockouts}{}
\providecommand{\overrideIEEEmargins}{}

\IEEEoverridecommandlockouts                              

\title{\LARGE \bf
Learning Whole-Body Kinetic Coordination for Wheeled-Quadruped Long-Range Throwing
}

\author{Qinglian Wu$^{1}$, Yidian Hu$^{1}$, Weiqi Fan$^{2}$, Yutian Zhang$^{3}$, Siyuan Ma$^{4}$, Jiaqi Zhai$^{5}$, Wei Dong$^{5,\dagger}$, and Li Xu$^{1,\dagger}$%
\thanks{$^{1}$Harbin Institute of Technology, Harbin, China.}%
\thanks{$^{2}$Hainan University, Haikou, China.}%
\thanks{$^{3}$Zhejiang University, Hangzhou, China.}%
\thanks{$^{4}$Tsinghua University, Beijing, China.}%
\thanks{$^{5}$Hangzhou Yunshenchu Technology Co., Ltd. (DEEP Robotics), Hangzhou, China.}%
\thanks{$^{\dagger}$Corresponding authors: Li Xu and Wei Dong.}%
}

\begin{document}

\maketitle
\thispagestyle{empty}
\pagestyle{empty}

\begin{abstract}

Long-range throwing on unstructured terrain is a key capability for mobile robots in delivery and retrieval tasks, but it requires high release velocity, coordinated whole-body motion, and robustness to terrain-induced disturbances. We present a whole-body kinetic-coordination framework for a wheeled-quadruped manipulator that combines kinetic-chain arm modeling with SQP-based trajectory optimization and a reference-guided reinforcement learning policy for coordinated base motion. The manipulator is first reduced to a kinetic-chain model and optimized under kinematic and dynamic constraints to produce a high-speed throwing reference. On a stationary base, the optimized arm trajectory reaches a release speed of 12.75~m/s and a throwing distance of 16.68~m. When whole-body coordination is enabled, the system further increases the release speed to 16.64~m/s and the throwing distance to 28.25~m. We then integrate the mobility of the wheeled-quadruped platform and validate the complete policy through hardware experiments on flat ground, grass, slopes, and rough roads. The results show that kinetic-chain optimization and whole-body coordination together enable robust long-range throwing across diverse unstructured terrains.

\end{abstract}

\section{INTRODUCTION}

Mobile robots with manipulators are increasingly used for outdoor delivery, inspection, and intervention tasks, where targets often lie beyond the immediate manipulation workspace. Throwing offers an efficient way to extend the reachable range, but many practical scenarios require the robot to release the payload from a moving platform rather than a stationary base. This run-and-throw setting turns throwing from an arm-centric skill into a whole-body mobile manipulation problem.

Executing high-speed throwing on unstructured terrain creates a tightly coupled whole-body control problem. The arm motion required for release velocity generates large self-induced reaction forces on the floating base, while uneven terrain continuously changes support conditions and body posture. The controller must therefore coordinate base motion and manipulator acceleration in a single dynamic process, rather than treating locomotion stabilization and throwing execution as separate modules.

Beyond stability, long-range throwing requires efficient kinetic energy transfer from proximal joints to the distal end-effector. Inspired by the human kinetic chain, which amplifies release velocity via proximal-to-distal motion propagation, the robot must preserve this inertial coupling despite base disturbances. This motivates a unified framework combining terrain-robust whole-body stabilization with kinetic-chain-guided arm motion.

Prior work has studied terrain-robust locomotion, whole-body loco-manipulation, and robotic throwing from complementary perspectives. Learning-based locomotion controllers have enabled legged robots to traverse irregular terrains through large-scale reinforcement learning with simulated-response and implicit-terrain modeling~\cite{long2024hybrid,nahrendra2023dreamwaq,lee2020learning}, but these methods mainly address base mobility and do not consider coordination with a mounted manipulator. Whole-body control enables legged and wheeled-legged robots to coordinate locomotion with manipulation~\cite{ma2022combining,fu2023deep,jiang2024learning}, yet these methods typically focus on general manipulation or tracking tasks rather than highly dynamic motions on unstructured outdoor terrains. Robotic throwing has been explored through kinetic-chain modeling~\cite{senoo2008kinetic} and learning-based planning~\cite{zeng2020tossingbot,liu2024tube,liu2022adaptive}, but most approaches are designed for fixed or slow-moving bases and do not address whole-body dynamic coordination. More recently, studies have begun integrating throwing with legged systems~\cite{werner2024dynamic,munn2024wholebody,ma2025accurate}, showing that base motion can improve throwing performance. Nevertheless, long-range throwing on unstructured terrain remains largely unexplored. In this setting, rapid arm acceleration can severely disturb a floating base, requiring coordinated kinetic-chain energy transfer with robust whole-body stabilization.

\begin{figure}[t]
  \centering
  \includegraphics[width=0.88\columnwidth]{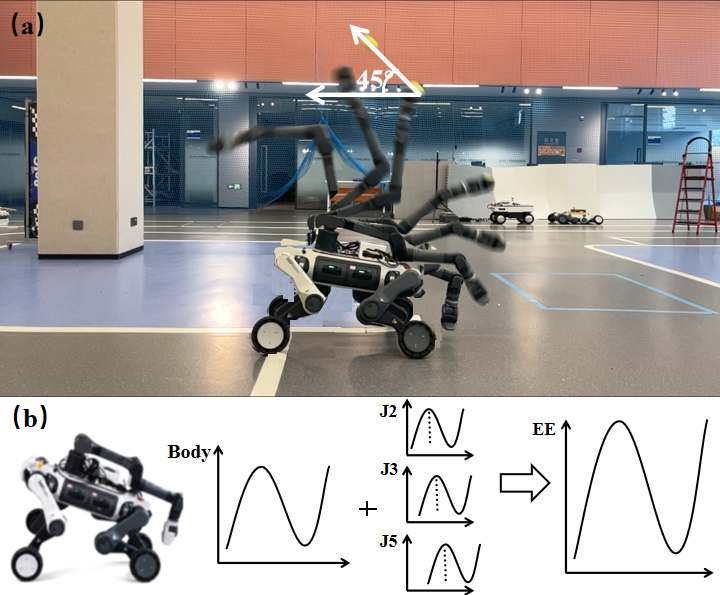}
  \caption{(a) High-speed throwing by the wheeled-quadruped robot with whole-body coordination. (b) Schematic of manipulator kinetic-chain energy transfer.}
  \label{fig:real_throwing}
\end{figure}

\begin{figure*}[t]
  \centering
  \includegraphics[width=0.88\textwidth]{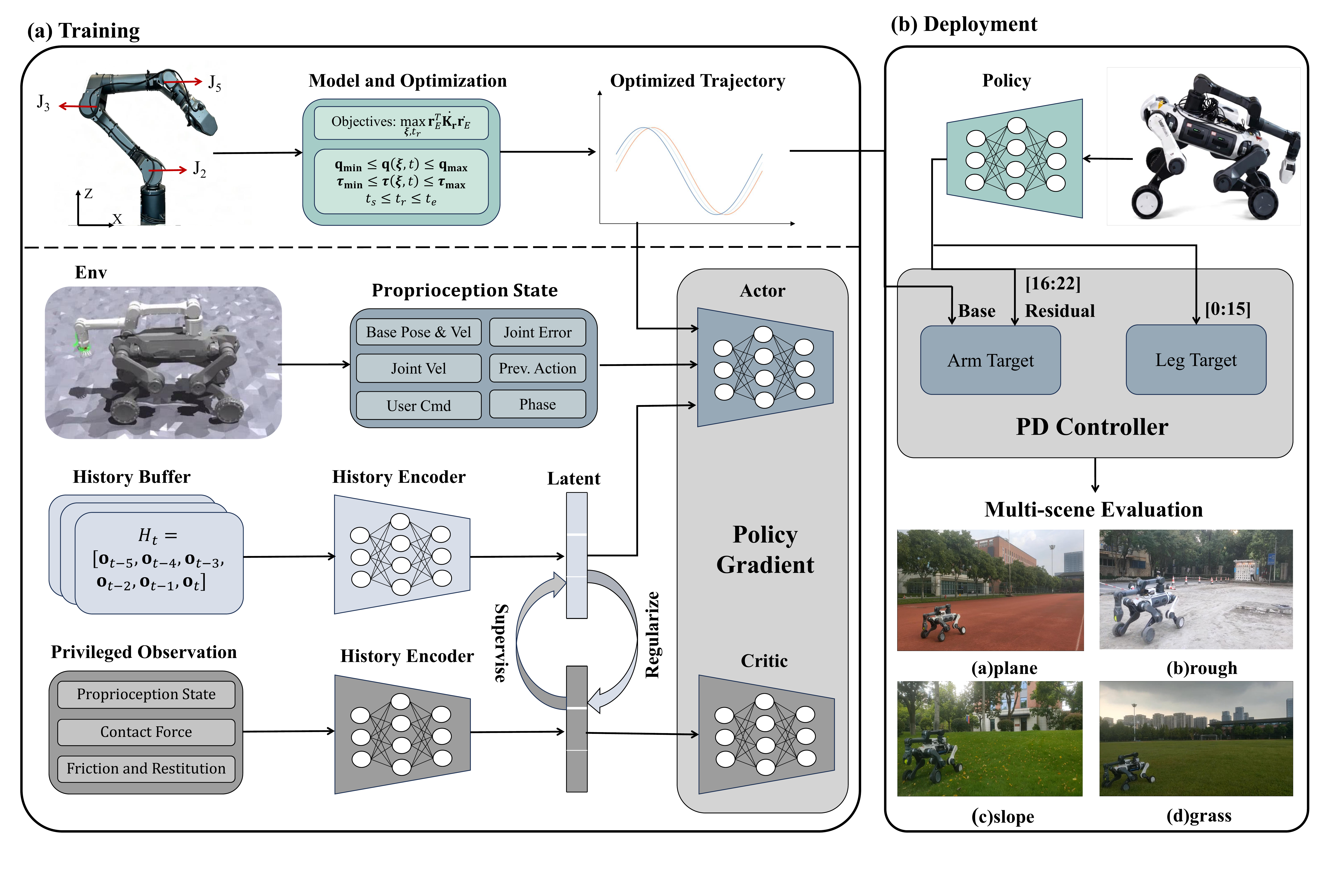}
  \vspace{-2.0em}
  \caption{Overview of the proposed control framework, which integrates arm trajectory optimization, whole-body reinforcement learning, and real-world deployment. The manipulator reference is optimized to enhance kinetic-chain energy transfer, while the wheeled-quadruped base is trained jointly in simulation with a history-based latent representation for terrain adaptation. The resulting policy enables terrain-robust long-range throwing and is evaluated across multiple real-world scenarios.}
  \label{fig:framework}
\end{figure*}

In this work, we address terrain-robust long-range throwing with a reference-guided whole-body reinforcement learning framework. We first optimize a high-speed arm reference trajectory with SQP on an approximate kinetic-chain model, then train a whole-body policy to track this reference while coordinating the manipulator and wheeled-quadruped base. The resulting formulation balances proximal-to-distal energy transfer with traction and whole-body stability. Experiments show that the unified policy enables consistent long-range throwing across diverse unstructured terrains.

We summarize our contributions as follows:
\begin{itemize}
  \item A whole-body RL framework that coordinates the wheeled-quadruped base and manipulator under terrain disturbances and arm-induced reaction forces.

  \item A kinetic-chain-based trajectory optimization method that increases release velocity through proximal-to-distal energy transfer under manipulator constraints.

  \item Extensive hardware and simulation validation of the proposed method on flat ground, rough roads, slopes, and grass, demonstrating robust run-and-throw behavior.
\end{itemize}

\section{Related Work}

Whole-body loco-manipulation has progressed from decoupled locomotion-manipulation pipelines to unified policies that coordinate the robot body and arm. Early methods combined a learned locomotion policy with model-based manipulation, allowing the base controller to compensate for arm-induced disturbances but still relying on a decoupled control structure~\cite{ma2022combining}, while later work learned unified policies that integrate locomotion and manipulation within a single control framework~\cite{fu2023deep}. Whole-body task-space tracking has also been improved through a reward-fusion module that enables precise 6D end-effector pose tracking~\cite{jiang2024learning}. Recent coordinated policies have achieved dynamic skills such as badminton playing and accurate prehensile throwing with legged manipulators~\cite{ma2025accurate,ma2025badminton}. In contrast, this work focuses on terrain-robust long-range throwing, where the robot must coordinate whole-body stability on unstructured terrain with explicit kinetic-chain energy transfer for high-speed release.

\textit{a) Unified Whole-Body Control:}
Whole-body control has been increasingly studied as an alternative to modular locomotion-manipulation pipelines. Fu et al.~\cite{fu2023deep} showed that a unified reinforcement learning policy can coordinate the legs and arm more effectively than decoupled controllers, highlighting the promise of unified control for mobile manipulation. However, existing methods have largely focused on relatively moderate manipulation motions and do not explicitly address complex terrain conditions, leaving the manipulator's full dynamic potential underexploited. Our work builds on this unified-policy perspective and targets terrain-robust long-range throwing by coordinating whole-body stability with kinetic-chain-driven high-speed release under terrain disturbances.

\textit{b) Kinetic-Chain-Based Throwing:}
Kinetic-chain modeling has been used to generate high-speed throwing motions by exploiting proximal-to-distal energy transfer. Senoo et al.~\cite{senoo2008kinetic} derived approximate swing dynamics and decomposed the throwing motion into basis functions, showing that inertial coupling can amplify distal joint velocity for fast release. However, their formulation was developed for a fixed-base manipulator. Here, we extend this principle to a wheeled-quadruped platform, where whole-body coordination enables the arm to exploit a moving base and thereby expand its effective throwing capability under real joint constraints.

\section{Method}

As shown in Fig.~\ref{fig:framework}, the proposed framework addresses terrain-robust dynamic throwing in two stages: kinetic-chain arm optimization and whole-body RL control. We first model the inertial coupling among the active throwing joints and parameterize the motion with sinusoidal basis functions. The trajectory parameters and release time are then optimized by SQP to maximize the weighted end-effector release velocity under physical constraints. The resulting reference is embedded into a whole-body policy that coordinates the wheeled-quadruped base and applies small residual corrections around the optimized arm reference trajectory.

\subsection{Kinetic-chain Model} 

We construct a reduced 3-DOF throwing model for the manipulator's kinetic chain. The model retains the active sagittal joints $J_2$, $J_3$, and $J_5$, which share parallel rotation axes along the local $Y$-axis and drive the forward throwing motion, while $J_1$, $J_4$, and $J_6$ are held fixed and absorbed into three equivalent rigid-body groups:
\begin{equation}
\begin{aligned}
    G_2 &= \{\mathrm{L}_2\}, \\
    G_3 &= \{\mathrm{L}_3, \mathrm{L}_4\}, \\
    G_5 &= \{\mathrm{L}_5, \mathrm{L}_6, \mathrm{EE}\}.
\end{aligned}
\end{equation}

We use the following reduced analysis to guide a physically motivated basis-function parameterization, with trajectory feasibility and torque limits evaluated later using the full robot dynamics. Gravity is omitted in this analytical step to isolate the inertial coupling that drives the kinetic-chain effect.
For each group, we define the equivalent mass, first mass moment, and rotational inertia as:
\begin{equation}
\begin{aligned}
    m_r &= \sum_{i\in G_r} m_i, \quad 
    \boldsymbol{\mu}_r = \sum_{i\in G_r} m_i\boldsymbol{\rho}_i^{(r)}, \\
    \bar{J}_r &= \sum_{i\in G_r} \left(I_{iy,i} + m_i\|\boldsymbol{\rho}_i^{(r)}\|^2\right),
\end{aligned}
\end{equation}
where $\boldsymbol{\rho}_i^{(r)} \in \mathbb{R}^2$ is the planar center-of-mass position of rigid body $i$ expressed in the local frame of group $G_r$. Let $\boldsymbol{a}$ and $\boldsymbol{b}$ denote the fixed planar offsets between adjacent active joint frames. The distal-mass-induced inertia terms are:
\begin{equation}
    B_{23} = (m_3 + m_5)\|\boldsymbol{a}\|^2, \quad 
    B_{35} = m_5\|\boldsymbol{b}\|^2.
\end{equation}
The configuration-dependent coupling terms are defined as:
\begin{equation}
\begin{aligned}
    A_{23}(q_3) &= \boldsymbol{a}^{T}\boldsymbol{R}(q_3)(\boldsymbol{\mu}_3 + m_5\boldsymbol{b}), \\
    A_{25}(q_3+q_5) &= \boldsymbol{a}^{T}\boldsymbol{R}(q_3+q_5)\boldsymbol{\mu}_5, \\
    A_{35}(q_5) &= \boldsymbol{b}^{T}\boldsymbol{R}(q_5)\boldsymbol{\mu}_5.
\end{aligned}
\end{equation}
With
\begin{equation}
\begin{aligned}
    \dot{\theta}_2 &= \dot{q}_2, \quad 
    \dot{\theta}_3 = \dot{q}_2 + \dot{q}_3, \\
    \dot{\theta}_5 &= \dot{q}_2 + \dot{q}_3 + \dot{q}_5,
\end{aligned}
\end{equation}
the total kinetic energy of the reduced planar chain is written as:
\begin{equation}
\begin{aligned}
    T &= \frac{1}{2}(\bar{J}_2 + B_{23})\dot{\theta}_2^2 
         + \frac{1}{2}(\bar{J}_3 + B_{35})\dot{\theta}_3^2 \\
      &\quad + \frac{1}{2}\bar{J}_5\dot{\theta}_5^2 
         + A_{23}\dot{\theta}_2\dot{\theta}_3 \\
      &\quad + A_{25}\dot{\theta}_2\dot{\theta}_5 
         + A_{35}\dot{\theta}_3\dot{\theta}_5.
\end{aligned}
\end{equation}

The inertia matrix is obtained from the Hessian of the kinetic energy:
\begin{equation}
    \mathbf{M}(\mathbf{q}) = \frac{\partial^2T}{\partial\dot{\mathbf{q}}\partial\dot{\mathbf{q}}^{T}},
\end{equation}
and can be expressed as:
\begin{equation}
    \mathbf{M}(\mathbf{q}) = 
    \begin{bmatrix}
        M_{22} & M_{23} & M_{25} \\
        M_{23} & M_{33} & M_{35} \\
        M_{25} & M_{35} & M_{55}
    \end{bmatrix},
\end{equation}
where
\begin{equation}
\begin{aligned}
    M_{22} &= \bar{J}_2 + B_{23} + \bar{J}_3 + B_{35} + \bar{J}_5 \\
           &\quad + 2A_{23} + 2A_{25} + 2A_{35}, \\
    M_{23} &= \bar{J}_3 + B_{35} + \bar{J}_5 + A_{23} + A_{25} + 2A_{35}, \\
    M_{25} &= \bar{J}_5 + A_{25} + A_{35}, \\
    M_{33} &= \bar{J}_3 + B_{35} + \bar{J}_5 + 2A_{35}, \\
    M_{35} &= \bar{J}_5 + A_{35}, \quad 
    M_{55} = \bar{J}_5.
\end{aligned}
\end{equation}
The reduced dynamics are then given by:
\begin{equation}
    \boldsymbol{\tau} = \mathbf{M}(\mathbf{q})\ddot{\mathbf{q}} + \mathbf{h}(\mathbf{q},\dot{\mathbf{q}}),
\end{equation}
where $\mathbf{h}(\mathbf{q},\dot{\mathbf{q}})$ contains the Coriolis and centrifugal terms.

Following the kinetic-chain basis-function decomposition, $J_2$ is treated as the proximal driving joint, while $J_3$ and $J_5$ are analyzed as distal coordinates driven mainly by inertial coupling. During release acceleration, we assume:
\begin{equation}
    |\dot{q}_2| \gg |\dot{q}_3|, |\dot{q}_5|, \quad 
    \dot{q}_2 \simeq \mathrm{const.}, \quad 
    \ddot{q}_2 \simeq 0,
\end{equation}
and
\begin{equation}
    \tau_3 \simeq 0, \quad \tau_5 \simeq 0.
\end{equation}
This zero-torque condition is a passive-response approximation used only to extract the coupling-induced basis functions; the optimized trajectory later allows all three joints to satisfy their actuator limits. The distal rows of the dynamics are therefore approximated as:
\begin{equation}
    \begin{bmatrix}
        M_{33} & M_{35} \\
        M_{35} & M_{55}
    \end{bmatrix}
    \begin{bmatrix}
        \ddot{q}_3 \\
        \ddot{q}_5
    \end{bmatrix}
    +
    \begin{bmatrix}
        h_3 \\
        h_5
    \end{bmatrix}
    \simeq
    \begin{bmatrix}
        0 \\
        0
    \end{bmatrix}.
\end{equation}

To obtain the basis function of $J_3$, we retain the dominant acceleration term of $J_3$ and treat the remaining distal coordinate as uniform motion:
\begin{equation}
    \ddot{\mathbf{q}} = \begin{bmatrix} 0 & \ddot{q}_3 & 0 \end{bmatrix}^{T}.
\end{equation}
Substituting this into the distal subsystem yields:
\begin{equation}
    (\bar{J}_3 + B_{35} + \bar{J}_5 + 2A_{35})\ddot{q}_3 - (A'_{23} + A'_{25})\dot{q}_2^2 \simeq 0.
\end{equation}
where $A'_{23}$ denotes the local derivative of $A_{23}$ with respect to $q_3$, and $A'_{25}$ denotes the derivative of $A_{25}$ with respect to its scalar argument $q_3+q_5$ under this projection. Let $q_3^{\ast}$ be a local equilibrium satisfying $(A'_{23}+A'_{25})|_{q_3=q_3^{\ast}}=0$. Around this equilibrium, for a locally restoring branch with $k_3^{\ast}>0$, the coupling term is approximated as:
\begin{equation}
    A'_{23} + A'_{25} \simeq -k_3^{\ast}\delta q_3,
\end{equation}
where $\delta q_3 = q_3 - q_3^{\ast}$. This gives:
\begin{equation}
    (\bar{J}_3 + B_{35} + \bar{J}_5 + 2A_{35})^{\ast}\delta\ddot{q}_3 + k_3^{\ast}\dot{q}_2^2\delta q_3 \simeq 0.
\end{equation}
Therefore, $J_3$ has a sinusoidal local solution:
\begin{equation}
    q_3(t) \simeq c_3 + \alpha_3\sin(\omega_3 t + \phi_3).
\end{equation}

Applying the same approximation to $J_5$ gives
\begin{equation}
    \bar{J}_5\delta\ddot{q}_5 + k_5^{\ast}\dot{q}_2^2\delta q_5 \simeq 0,
\end{equation}
where $\delta q_5 = q_5 - q_5^{\ast}$ and $k_5^{\ast} > 0$. Therefore,
\begin{equation}
    q_5(t) \simeq c_5 + \alpha_5\sin(\omega_5 t + \phi_5).
\end{equation}

For computational convenience, the proximal joint $J_2$ is also parameterized by a sinusoidal basis function. The approximate analysis determines the functional form, while the amplitudes, frequencies, phases, and offsets are optimized later under robot constraints. The reduced throwing trajectory is therefore written as:
\begin{equation}
    \mathbf{q}(t) = 
    \begin{bmatrix}
        c_2 + \alpha_2\sin(\omega_2 t + \phi_2) \\
        c_3 + \alpha_3\sin(\omega_3 t + \phi_3) \\
        c_5 + \alpha_5\sin(\omega_5 t + \phi_5)
    \end{bmatrix}.
\end{equation}

\subsection{Trajectory Optimization}

The kinetic-chain analysis above provides the functional form of the throwing motion, while the actual offsets, amplitudes, frequencies, phases, and release time are determined by constrained optimization. Following this basis-function strategy, the reduced joint trajectory is parameterized by:
\begin{equation}
\begin{aligned}
    \boldsymbol{\xi} &= [c_2, \alpha_2, \omega_2, \phi_2, c_3, \alpha_3, \omega_3, \phi_3, \\
                     &\qquad c_5, \alpha_5, \omega_5, \phi_5]^T.
\end{aligned}
\end{equation}
Given $\boldsymbol{\xi}$ and the release time $t_r$, the end-effector velocity is computed from the forward kinematics as:
\begin{equation}
    \dot{\mathbf{r}}_E = \mathbf{J}_E(\mathbf{q}(\boldsymbol{\xi},t_r))\dot{\mathbf{q}}(\boldsymbol{\xi},t_r) \triangleq f(\boldsymbol{\xi},t_r),
\end{equation}
where $\mathbf{J}_E$ is the end-effector Jacobian.

The active throwing interval $[t_s,t_e]$ is defined from the zero-velocity points of the sinusoidal joint references, and the release time is constrained within this interval. We optimize the trajectory parameters and release time to maximize the weighted end-effector release speed:
\begin{equation}
\begin{aligned}
    \max_{\boldsymbol{\xi},\,t_r} \quad & \dot{\mathbf{r}}_E^{T}(t_r) \mathbf{K}_r \dot{\mathbf{r}}_E(t_r) \\
    \text{s.t.} \quad & \mathbf{q}_{\min} \le \mathbf{q}(\boldsymbol{\xi},t) \le \mathbf{q}_{\max}, \\
                      & \dot{\mathbf{q}}_{\min} \le \dot{\mathbf{q}}(\boldsymbol{\xi},t) \le \dot{\mathbf{q}}_{\max}, \\
                      & \boldsymbol{\tau}_{\min} \le \boldsymbol{\tau}(\boldsymbol{\xi},t) \le \boldsymbol{\tau}_{\max}, \\
                      & \forall t \in [t_s,t_r], \\
                      & t_s \le t_r \le t_e.
\end{aligned}
\label{eq:optimization}
\end{equation}

Here, $\mathbf{K}_r$ is a positive definite weighting matrix that emphasizes the release-velocity components aligned with the desired throwing direction. The joint torque is evaluated using the full robot dynamics:
\begin{equation}
    \boldsymbol{\tau}(\boldsymbol{\xi},t) = \mathbf{M}(\mathbf{q})\ddot{\mathbf{q}} + \mathbf{h}(\mathbf{q},\dot{\mathbf{q}}) + \mathbf{g}(\mathbf{q}),
\end{equation}
so that the optimized trajectory respects the actual torque limits of the manipulator. When $\mathbf{K}_r = \frac{1}{2}m_e\mathbf{I}$, where $m_e$ is the effective end-effector mass, the objective is equivalent to maximizing the translational kinetic energy at release. In practice, this nonlinear constrained problem is solved via sequential quadratic programming (SQP), with the inequality constraints enforced over the active throwing interval $[t_s,t_r]$.

\subsection{Whole-Body Coordinated Control}

As illustrated in Fig.~\ref{fig:framework}, the optimized kinetic-chain trajectory is provided as a time-varying arm reference to the whole-body policy, which also receives the proprioception observation and history-encoded latent variable as inputs. The policy generates coordinated leg-wheel targets while applying small residual corrections to the optimized arm reference. Because the latent variable is inferred from proprioceptive history, the deployed policy does not require privileged information.

\begin{figure*}[!t]
  \centering
  \includegraphics[width=0.98\textwidth]{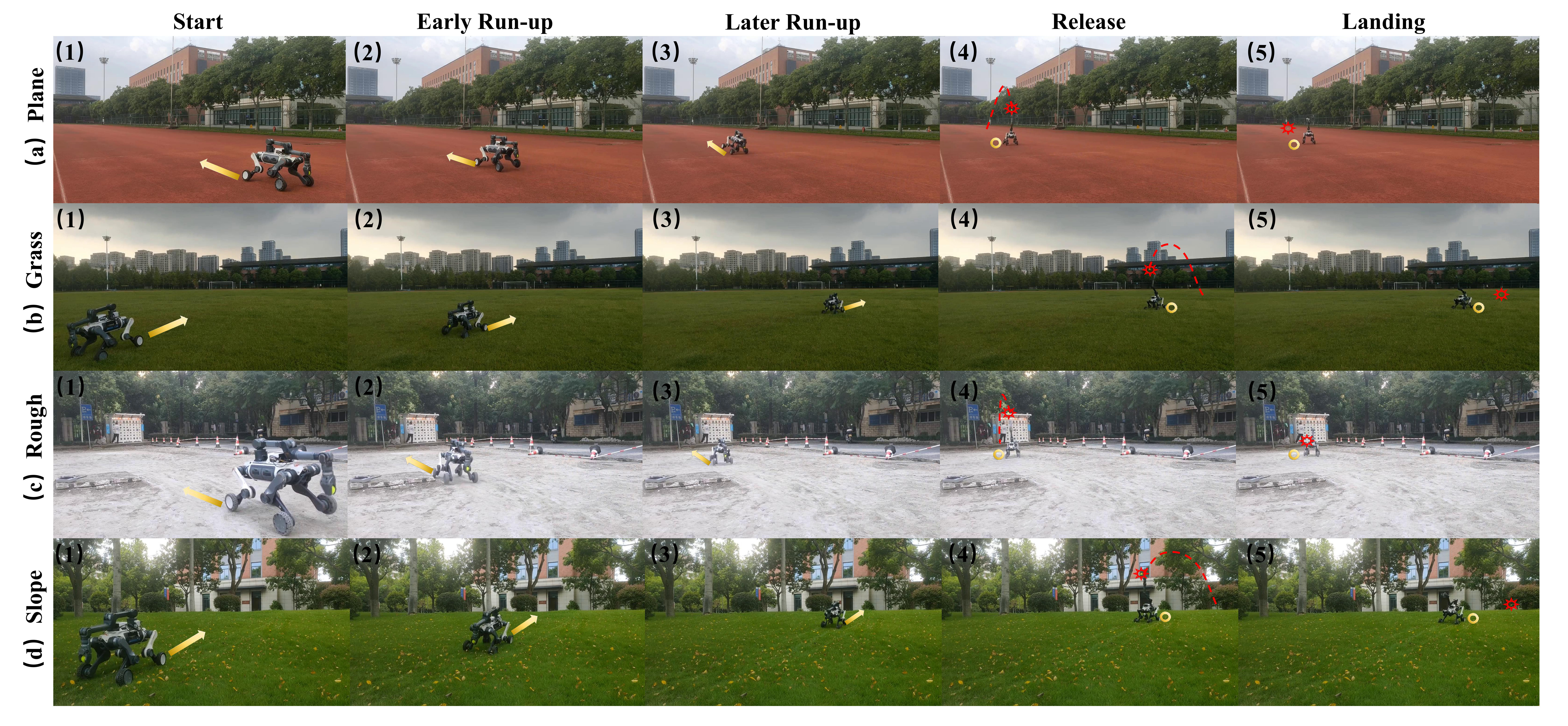}
  \vspace{-0.8em}
  \caption{Temporal phase sequence of the throwing motion across four representative terrains: (a) flat plane, (b) grass, (c) rough gravel road, and (d) grassy slope. Each row shows five key stages of the run-and-throw process: Start, Early Run-up, Late Run-up, Release, and Landing.}
  \label{fig:time_sequence}
\end{figure*}

\setcounter{table}{0}
\begin{table}[t]
    \centering
    \caption{Reward terms used in training.}
    \label{tab:reward_terms}
    \scriptsize
    \setlength{\tabcolsep}{2pt}
    \renewcommand{\arraystretch}{1.2}
    \begin{tabular}{p{2.1cm} p{5.3cm} c}
        \toprule
        Term & Implementation & Wt. \\
        \midrule
        tracking\_lin\_vel & $\exp\!\big({-}\|\mathbf{v}_{xy}^{\mathrm{cmd}} - \mathbf{v}_{xy}\|^2 / \sigma_{\mathrm{tr}}\big)$ & 1.00 \\
        tracking\_ang\_vel & $\exp\!\big({-}(\omega_z^{\mathrm{cmd}} - \omega_z)^2 / \sigma_{\mathrm{tr}}\big)$ & 0.50 \\
        lin\_vel\_z & $v_z^2$ & -0.25 \\
        orientation & $\|\mathbf{g}_{\mathrm{proj},xy}\|^2$ & -0.5 \\
        action\_rate & $\|\mathbf{a}_t - \mathbf{a}_{t-1}\|^2$ & -0.01 \\
        hip\_pos & $\|\mathbf{q}_{\mathrm{hip}} - \mathbf{q}_{\mathrm{hip}}^{\mathrm{def}}\|^2$ & -0.8 \\
        throw\_velocity & $g_{\mathrm{rel}} \, v_{\mathrm{tar}} \tanh\!\big(\|\dot{\mathbf{r}}_E\| / v_{\mathrm{tar}}\big)$ & 3.20 \\
        arm\_tracking & $s_{\mathrm{ph}} \exp\!\big({-}\|W(\mathbf{q}_{\mathrm{arm}}^{\mathrm{ref}} - \mathbf{q}_{\mathrm{arm}})\|^2 / (2\sigma_{\mathrm{arm}}^2)\big)$ & 0.75 \\
        base\_throw\_coord & $\mathbb{I}_{\mathrm{load}} f(e_{\mathrm{ld}}) + \mathbb{I}_{\mathrm{rel}} f(e_{\mathrm{re}}) + \mathbb{I}_{\mathrm{rec}} f(e_{\mathrm{rc}})$ & 1.10 \\
        \bottomrule
    \end{tabular}
\end{table}

\setcounter{table}{1}
\begin{table}[t]
    \centering
    \caption{Domain randomization used in training.}
    \label{tab:domain_rand}
    \scriptsize
    \renewcommand{\arraystretch}{1.1}
    \begin{tabular}{ll}
        \toprule
        Item & Setting \\
        \midrule
        Friction coefficient & $[0.2, \, 1.5]$ \\
        Added base mass & $[-2, \, 6]$ kg \\
        Base COM offset ($x,y,z$) & $[-0.05, \, 0.05]$ m \\
        Added gripper mass & $[0, \, 0.5]$ kg \\
        PD stiffness scale & $[0.9, \, 1.1]$ \\
        PD damping scale & $[0.9, \, 1.1]$ \\
        Action latency & $0$--$5$ control steps \\
        Push interval & 8 s \\
        Max push velocity & 0.2 m/s \\
        \bottomrule
    \end{tabular}
\end{table}

\subsubsection{Reference-Guided Whole-Body Policy}

The manipulator follows the optimized throwing motion through a time-parameterized joint-space reference trajectory. The reference arm posture is written as:
\begin{equation}
    \mathbf{q}_{\mathrm{arm}}^{\mathrm{ref}}(t) =
    \begin{bmatrix}
        q_1^h & q_2^{\mathrm{ref}}(t) & q_3^{\mathrm{ref}}(t) & q_4^h & q_5^{\mathrm{ref}}(t) & q_6^h
    \end{bmatrix}^{T}
\end{equation}
where the time-varying terms are obtained from the optimized sinusoidal trajectory.

The policy input is organized as a proprioception observation vector, which collects internal sensor information, command information, reference-phase information, and last action required for coordinating the leg-wheel system with the throwing arm. The current observation is defined as:
\begin{equation}
    \mathbf{o}_t =
    \begin{bmatrix}
        \mathbf{s}_t^{\mathrm{base}} & \Delta\mathbf{q}_t & \dot{\mathbf{q}}_t & \mathbf{a}_{t-1} & \mathbf{c}_t & \mathbf{p}_t & \mathbf{e}_t^{\mathrm{arm}}
    \end{bmatrix}^{T}
    \in \mathbb{R}^{73}
\end{equation}
where $\mathbf{s}_t^{\mathrm{base}} \in \mathbb{R}^{5}$ contains the base roll, pitch, and angular velocity, $\Delta\mathbf{q}_t \in \mathbb{R}^{18}$ denotes non-wheel joint position deviations, $\dot{\mathbf{q}}_t \in \mathbb{R}^{22}$ denotes joint velocities, $\mathbf{a}_{t-1} \in \mathbb{R}^{19}$ is the previous action, $\mathbf{c}_t \in \mathbb{R}^{3}$ is the base velocity command, and $\mathbf{e}_t^{\mathrm{arm}} \in \mathbb{R}^{3}$ is the tracking error between the current and reference trajectories of joints $J_2$, $J_3$, and $J_5$. The reference-phase cue is:
\begin{equation}
    \mathbf{p}_t =
    \begin{bmatrix}
        \cos(2\pi\phi_t) & \sin(2\pi\phi_t) & m_t
    \end{bmatrix}^{T}
\end{equation}
where $\phi_t$ denotes the normalized throwing phase and $m_t$ marks the active throwing interval, helping the policy separate preparation, release, and recovery stages.

Together with the history latent variable introduced in the next subsection, the whole-body policy is written as:
\begin{equation}
    \mathbf{a}_t = \pi_{\theta}(\mathbf{o}_t, \mathbf{z}_t^{h}).
\end{equation}
In the current implementation, $\mathbf{a}_t \in \mathbb{R}^{19}$ comprises twelve leg actions, four wheel actions, and three residual arm actions for joints $J_2$, $J_3$, and $J_5$. Since the arm reference is optimized under a nominal stationary-base assumption, it may not remain optimal when executed on the moving wheeled-quadruped base during RL training. The residual arm actions therefore compensate the optimized reference, and the resulting arm targets are sent together with the leg-wheel targets to the low-level PD controllers.

\setcounter{table}{2}
\begin{table*}[!t]
    \centering
    \caption{Ablation results for task performance, kinetic-chain contribution, and whole-body coordination. Best denotes the best trial among all evaluated cycles, and Mean denotes the average over evaluated cycles.}
    \label{tab:ablation_results}
    \scriptsize
    \setlength{\tabcolsep}{2.6pt}
    \renewcommand{\arraystretch}{1.15}
    \resizebox{\textwidth}{!}{%
    \begin{tabular}{llcccc@{\hspace{0.6em}\vrule width 0.35pt\hspace{0.6em}}ccc@{\hspace{0.6em}\vrule width 0.35pt\hspace{0.6em}}ccc}
        \toprule
        \multirow{2}{*}{Stat.} &
        \multirow{2}{*}{Variant} &
        \multicolumn{4}{c@{\hspace{0.6em}\vrule width 0.35pt\hspace{0.6em}}}{\textbf{Task-level throwing performance}} &
        \multicolumn{3}{c@{\hspace{0.6em}\vrule width 0.35pt\hspace{0.6em}}}{\textbf{Kinetic-chain contribution}} &
        \multicolumn{3}{c}{\textbf{Whole-body coordination contribution}} \\
        \cmidrule(l{0.4em}r{0.9em}){3-6}
        \cmidrule(l{0.9em}r{0.9em}){7-9}
        \cmidrule(l{0.9em}r{0.4em}){10-12}
        &
        &
        Distance (m) &
        Release speed (m/s) &
        Angle (rad) &
        Height (m) &
        Distal vel. gain &
        Timing span (s) &
        EE speed (m/s) &
        Pitch vel. (rad/s) &
        Ang. momentum (N$\cdot$m$\cdot$s) &
        Work (J) \\
        \midrule
        \multirow{4}{*}{Best}
        & None
        & 3.88 & 5.99 & 0.21 & 1.34
        & 0.90 & 0.88 & 8.16
        & 0.76 & 0.41 & 6.24 \\
        & KC
        & 16.68 & 12.76 & 0.56 & 1.25
        & 0.99 & 1.68 & 15.86
        & 1.38 & 0.74 & 7.17 \\
        & WBC
        & 6.46 & 8.54 & 0.25 & 1.36
        & 1.85 & 1.08 & 10.90
        & 4.46 & 2.38 & 89.76 \\
        & \textbf{KC+WBC}
        & \textbf{28.25} & \textbf{16.65} & \textbf{0.62} & \textbf{1.20}
        & \textbf{0.97} & \textbf{1.66} & \textbf{18.93}
        & \textbf{8.80} & \textbf{4.70} & \textbf{193.13} \\
        \midrule
        \multirow{4}{*}{Mean}
        & None
        & $3.49 \pm 0.11$ & $5.96 \pm 0.32$ & $0.09 \pm 0.05$ & $1.38 \pm 0.04$
        & $0.82 \pm 0.07$ & $0.24 \pm 0.17$ & $6.99 \pm 0.62$
        & $0.15 \pm 0.08$ & $0.08 \pm 0.04$ & $1.49 \pm 0.78$ \\
        & KC
        & $9.75 \pm 0.43$ & $9.58 \pm 0.42$ & $0.50 \pm 0.05$ & $1.19 \pm 0.01$
        & $0.78 \pm 0.21$ & $0.19 \pm 0.17$ & $12.22 \pm 0.85$
        & $0.22 \pm 0.11$ & $0.12 \pm 0.06$ & $2.09 \pm 1.38$ \\
        & WBC
        & $5.28 \pm 0.11$ & $7.49 \pm 0.32$ & $0.22 \pm 0.05$ & $1.38 \pm 0.04$
        & $1.22 \pm 0.03$ & $0.17 \pm 0.17$ & $8.98 \pm 0.62$
        & $3.61 \pm 0.28$ & $1.93 \pm 0.15$ & $63.57 \pm 9.83$ \\
        & \textbf{KC+WBC}
        & \textbf{17.19} $\pm$ \textbf{2.42} & \textbf{12.64} $\pm$ \textbf{1.28} & \textbf{0.64} $\pm$ \textbf{0.09} & \textbf{1.16} $\pm$ \textbf{0.03}
        & \textbf{0.72} $\pm$ \textbf{0.07} & \textbf{0.17} $\pm$ \textbf{0.11} & \textbf{14.11} $\pm$ \textbf{1.76}
        & \textbf{7.43} $\pm$ \textbf{1.72} & \textbf{3.97} $\pm$ \textbf{0.92} & \textbf{128.93} $\pm$ \textbf{32.77} \\
        \bottomrule
\end{tabular}%
    }
\end{table*}

\subsubsection{History-Guided Latent Adaptation}

High-speed throwing on uneven terrain is partially observable because terrain properties, wheel-ground traction, and arm-induced reactions evolve over time and cannot be fully captured by instantaneous proprioception. Recent changes in posture, joint motion, wheel response, and tracking error therefore provide useful context for inference. We therefore introduce a history-dependent latent variable to infer the disturbance-sensitive state from recent proprioceptive observations:
\begin{equation}
    \mathbf{z}_t^{h} = E_h(\mathcal{H}_t), \qquad \mathcal{H}_t = \begin{bmatrix} \mathbf{o}_{t-K+1}, \ldots, \mathbf{o}_t \end{bmatrix},
\end{equation}
where $E_h$ is a history encoder applied to the recent proprioceptive sequence. In our implementation, the history length is $K=10$ and the latent dimension is $20$.

During training, we additionally use privileged simulation information $\mathbf{o}_t^{p}$, including height-map information, foot-contact information, and external disturbances, to construct a privileged latent representation:
\begin{equation}
    \mathbf{z}_t^{p} = E_p(\mathbf{o}_t^{p}),
\end{equation}
where $E_p$ is a two-layer multi-layer perceptron (MLP) with the same output dimension as $E_h$. As shown in Fig.~\ref{fig:framework}, this forms an asymmetric latent-training scheme: the history encoder provides the deployable latent input to the actor, while the privileged encoder provides a simulation-only latent input to the critic during training.

To align the history-based and privileged latent spaces, we use:
\begin{equation}
\begin{aligned}
    \mathcal{L}_{\mathrm{priv}} &= \left\| \mathbf{z}_t^{p} - \mathrm{sg}(\mathbf{z}_t^{h}) \right\|_2, \\
    \mathcal{L}_{\mathrm{hist}} &= \left\| \mathrm{sg}(\mathbf{z}_t^{p}) - \mathbf{z}_t^{h} \right\|_2,
\end{aligned}
\end{equation}
where $\mathrm{sg}(\cdot)$ denotes the stop-gradient operator. The stop-gradient operator separates the two update directions, allowing the privileged latent to regularize the history latent without collapsing both representations simultaneously. In practice, Proximal Policy Optimization (PPO)~\cite{schulman2017ppo} updates are alternated with periodic supervised history adaptation every 20 iterations, and the privileged regularization coefficient is increased gradually during training. At deployment, only the current observation and the history latent $\mathbf{z}_t^{h}$ are used.

\subsection{Training and Implementation}

We train the whole-body coordination policy with PPO using 4096 parallel environments. Each rollout contains 24 control steps, with a fixed learning rate of $2 \times 10^{-4}$, discount factor $\gamma=0.99$, and GAE parameter $\lambda=0.95$. Training runs for 8000 iterations.

The per-step reward is written as:
\begin{equation}
    r_t = \sum_k w_k r_{k,t}.
\end{equation}
The exact implementation of each reward component is detailed in Table~\ref{tab:reward_terms}. Here, $g_{\mathrm{rel}}$ is the release-time gate, $s_{\mathrm{ph}}$ is the phase-dependent arm-tracking scale, and $\mathbb{I}_{\mathrm{load}}$, $\mathbb{I}_{\mathrm{rel}}$, and $\mathbb{I}_{\mathrm{rec}}$ denote the load-up, release, and recovery windows, respectively.

To improve robustness under high-speed whole-body motion across different terrains, we apply compact domain randomization over contact, actuation, and disturbance parameters, as summarized in Table~\ref{tab:domain_rand}.

All experiments are conducted on a wheeled-quadruped manipulator composed of a DeepRobotics M20 Pro mobile base and a custom CM1 6-DoF arm. The nominal total mass of the platform is 41.43~kg. For the active throwing joints $J_2$, $J_3$, and $J_5$, the joint limits are $q_2 \in [-\pi, \, 0]$, $q_3 \in [0, \, \pi]$, and $q_5 \in [-\pi/2, \, \pi/2]$~rad, with effort limits of 130, 130, and 50~N$\cdot$m, respectively. The throwing object used in all subsequent experiments is a standard tennis ball.

\section{Experiments and Results}

We evaluate the proposed framework from three aspects. First, we analyze the SQP-optimized throwing reference and the resulting proximal-to-distal kinetic energy transfer. Second, we conduct simulation experiments and ablation studies to validate the effectiveness of the kinetic-chain reference and whole-body coordination. Third, we perform hardware experiments on different terrains to evaluate the robustness of the learned policy.

\subsection{Kinetic-Chain Validation}
\begin{figure}[!t]
    \centering
    \includegraphics[width=\linewidth]{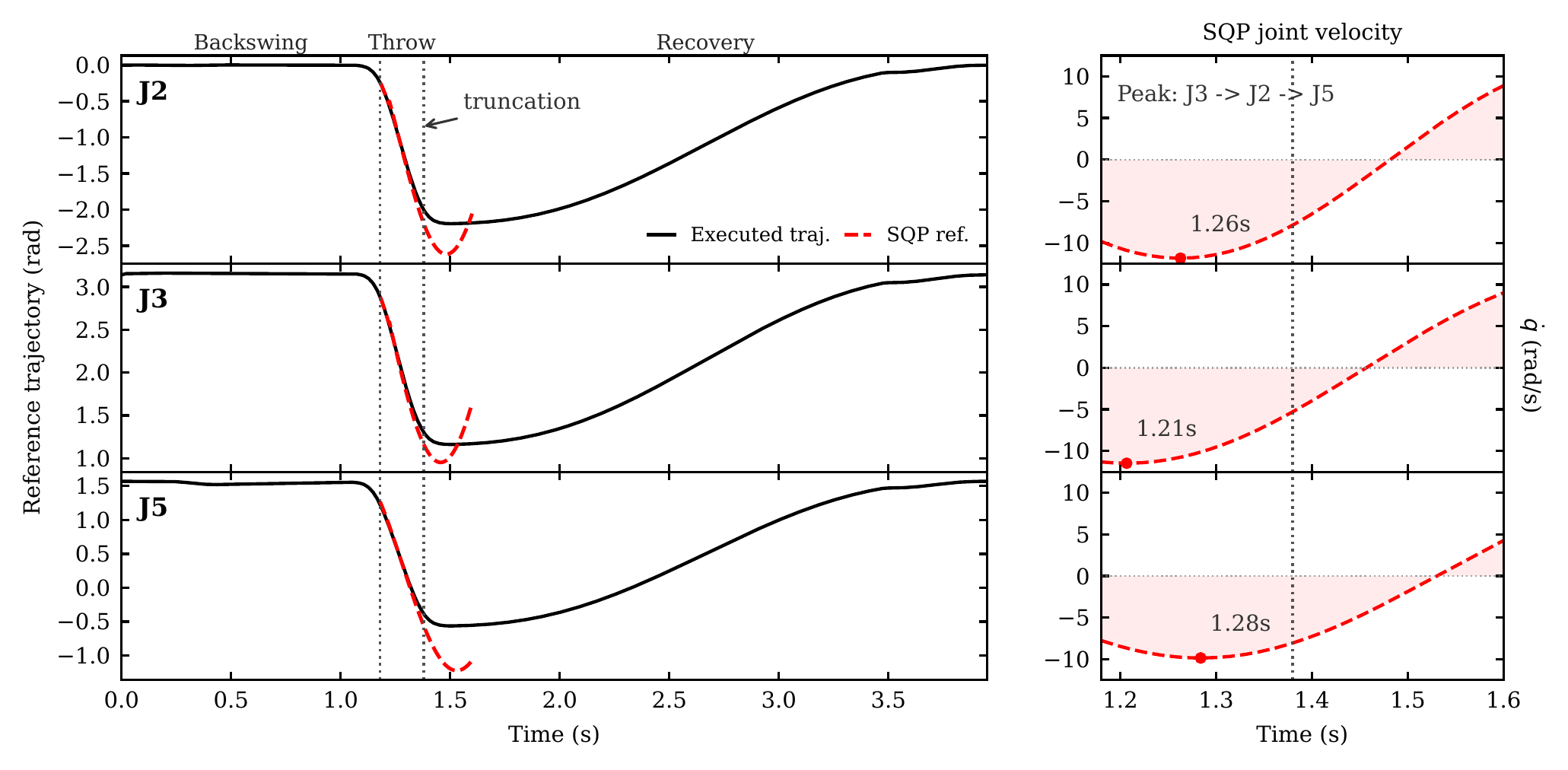}
    \caption{Optimized SQP trajectory for the active throwing joints.}
    \label{fig:sqp_trajectory}
\end{figure}

Fig.~\ref{fig:sqp_trajectory} shows the SQP-optimized arm references and the full-cycle controller commands for the active throwing joints $J_2$, $J_3$, and $J_5$. The full-cycle motion is organized into backswing, throwing, and recovery phases. The backswing phase allows the base to lean backward and prepare whole-body momentum before release. During the throwing phase, the controller follows the SQP reference to drive the arm rapidly toward the release configuration. After reaching the velocity peak, the optimized throwing segment is truncated and connected to a smooth recovery motion, preventing unnecessary high-speed arm continuation.

The velocity profiles reveal a staggered peak-timing pattern rather than a synchronous joint swing. Around the release interval, the peak velocities appear sequentially, with $J_3$, $J_2$, and $J_5$ reaching their peaks at approximately $1.21$~s, $1.26$~s, and $1.28$~s, respectively. This sequence follows an upstream-to-distal pattern, although not strictly in joint-index order, with the distal joint $J_5$ reaching its peak last. This staggered timing is consistent with the kinetic-chain objective of accumulating motion through the arm and increasing the distal release velocity. This reference serves as the nominal throwing motion for subsequent whole-body policy evaluation.

\subsection{Simulation Results}
\subsubsection{Simulation Setup and Metrics}
All four ablation variants are trained within the same reinforcement learning policy framework, with only the kinetic-chain reference and whole-body coordination components enabled or disabled across variants.
We compare four ablation variants:
\begin{itemize}
    \item \textbf{None}: without kinetic-chain reference and whole-body coordination.
    \item \textbf{KC}: with kinetic-chain reference only.
    \item \textbf{WBC}: with whole-body coordination only.
    \item \textbf{KC+WBC}: with both kinetic-chain reference and whole-body coordination.
\end{itemize}
Each method is evaluated over 100 environments with 10 throwing cycles per environment. Table~\ref{tab:ablation_results} reports the mean metrics over all cycles and the best trial for each variant.

Task-level metrics include throwing distance, release speed, release angle, and release height, where distance is computed from the measured release state using projectile motion. Kinetic-chain contribution is measured by distal velocity gain, joint peak timing span, and peak end-effector speed:
\begin{equation}
\begin{aligned}
    G_v &=
    \frac{\|\dot{\mathbf{r}}_E(t_r)\|}
    {\max_{i\in\{2,3,5\}}|\dot{q}_i(t_r)|}, \\
    \Delta t_{\mathrm{KC}}
    &=
    \max_i t_i^{\mathrm{peak}}
    -
    \min_i t_i^{\mathrm{peak}},
\end{aligned}
\end{equation}
where $t_i^{\mathrm{peak}}=\arg\max_t|\dot q_i(t)|$, and peak end-effector speed is the maximum speed over the throwing motion rather than the release speed. Whole-body coordination is measured by peak base pitch velocity, pitch angular momentum, and positive pitch work:
\begin{equation}
\begin{aligned}
    L_{b,p}(t)&=I_{b,yy}\omega_{b,p}(t), \\
    W_{b,p}^{+}
    &=
    \int_{t_0}^{t_r}
    \max(\tau_{b,p}(t)\omega_{b,p}(t),0)\,dt,
\end{aligned}
\end{equation}
where $\omega_{b,p}$ is the base pitch angular velocity, $\tau_{b,p}$ is the contact-induced pitch torque, $I_{b,yy}=0.533816~\mathrm{kg\,m^2}$, and $t_0$ is the first instant before release at which the base pitch velocity becomes positive.

\begin{figure}[t]
    \centering
    \includegraphics[width=0.73\columnwidth]{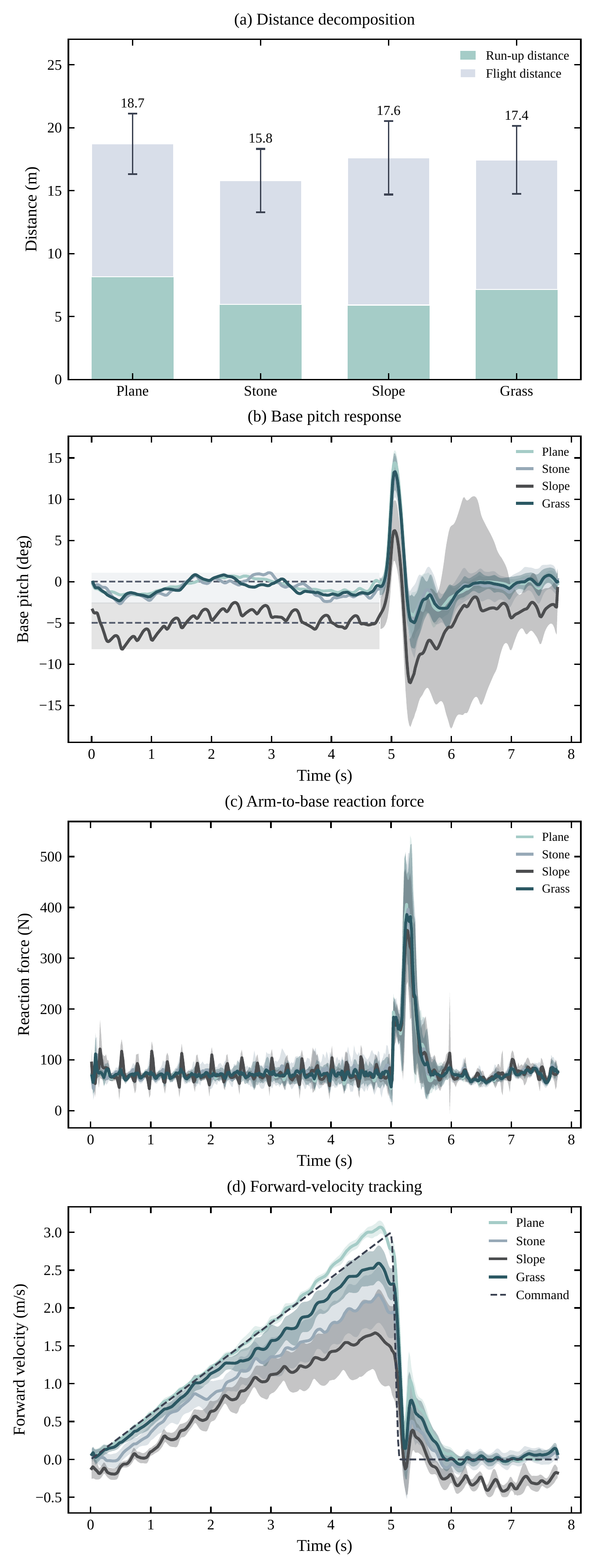}
    \caption{Hardware deployment results across flat ground, rough road, slope, and grass. (a) Throwing-distance decomposition. (b) Base pitch response. (c) Arm-to-base reaction force. (d) Forward-velocity tracking.}
    \label{fig:deployment}
\end{figure}

\subsubsection{Task-Level Throwing Performance}

The task-level results show that the kinetic-chain reference is the main source of performance gain, while whole-body coordination provides an additional boost. In isolation, KC raises the mean distance from 3.49~m to 9.75~m, and WBC increases it to 5.28~m, indicating complementary contributions to throwing range. When both components are enabled, KC+WBC achieves the best result, with a mean distance of 17.19~m and a mean release speed of 12.64~m/s. Table~III further shows that KC+WBC reaches a release angle of 0.64~rad but a lower release height than the None baseline, suggesting that release speed is the dominant factor when the launch angle and release height are already within a suitable range.

\subsubsection{Kinetic-Chain Contribution}
\par The results confirm that the kinetic-chain reference improves proximal-to-distal energy transfer, as reflected by substantial increases in both peak end-effector speed and release speed under both control settings. Although the distal velocity gain does not monotonically increase in all cases, this is not contradictory: the KC reference also raises the upstream joint speeds, so a lower normalized gain still corresponds to a faster overall release motion. More importantly, the optimized reference yields a shorter mean joint peak timing span than the non-kinetic baseline under both control settings, suggesting that the active joints follow a more compact and coordinated release profile. Together, these results show that KC improves both the magnitude and the temporal organization of the throwing motion, which is essential for high-speed long-range release.
\subsubsection{Whole-Body Coordination Contribution}

\par The whole-body coordination results show that the learned policy actively engages the wheeled-quadruped body during the throwing phase. Compared with the corresponding WBC-disabled variants, enabling WBC substantially increases peak base pitch velocity, peak pitch angular momentum, and positive pitch work. In particular, WBC raises the mean base pitch velocity from 0.15 to 3.61~rad/s and the mean positive pitch work from 1.49 to 63.57~J, while KC+WBC further increases them to 7.43~rad/s and 128.93~J. These gains indicate that the policy generates an active pitching motion that contributes additional proximal kinetic energy to the throw. Together with the task-level improvements in release speed and throwing distance, they show that WBC amplifies the throwing motion through coordinated body dynamics.

\subsection{Real-World Results}

\par We evaluate the complete KC / WBC policy through run-and-throw experiments on four representative terrains: a flat plane, a rough road, a slope, and grass. Unlike stationary throwing, each trial requires the robot to accelerate forward before release, coordinate the base and manipulator during the throwing motion, and recover its posture after payload release. For each terrain, we evaluate 50 run-and-throw cycles and analyze the resulting travel distance, base pitch response, arm-to-base reaction force, and forward-velocity tracking.

\subsubsection{Run-and-Throw Performance}

\par The robot completed the run-and-throw sequence on all four terrains. As shown in Fig.~\ref{fig:deployment}(a), the mean total distances were 18.72~m on the plane, 15.80~m on the rough road, 17.62~m on the slope, and 17.45~m on grass, decomposed into run-up and flight components. The run-up phase is intentionally conservative to preserve stability before release, so part of the potential flight gain is traded off for safer execution; nevertheless, the moving-base strategy still substantially extends the overall range.

\subsubsection{Whole-Body Dynamic Response}

\par Figure~5(b) shows the base pitch response throughout the run-and-throw cycle. During the run-up phase, the base pitch remains close to the terrain-dependent reference, with the three approximately level terrains centered near $0^\circ$ and the slope centered near $-5^\circ$. At the onset of throwing, the base exhibits a rapid forward pitching motion synchronized with the arm acceleration, reaching mean peak values of approximately $14.4^\circ$, $12.7^\circ$, $6.2^\circ$, and $13.3^\circ$ on the plane, rough road, slope, and grass, respectively. After release, the base undergoes only a brief backward recoil and quickly returns to its terrain-dependent steady posture, indicating fast post-release recovery.

\par The arm-to-base reaction force in Fig.~5(c) further reveals the dynamic interaction between the manipulator and the mobile platform. The reaction force remains moderate during the run-up phase and increases sharply around the throwing interval, peaking at approximately $406$~N. Despite this large impulse, the base remains stable and recovers promptly, showing that the wheeled-quadruped platform can absorb arm-induced disturbances without losing balance. This response also reflects effective proximal-to-distal energy transfer through the kinetic chain, which contributes to the high-speed release.

\subsubsection{Velocity Tracking and Recovery}

\par Figure~\ref{fig:deployment}(d) highlights the temporal coordination of the run-and-throw policy. During the run-up phase, the robot tracks the forward-speed command while accelerating toward the release window, reaching peak mean velocities of approximately $3.07$~m/s on the plane, $2.17$~m/s on the stone road, $1.67$~m/s on the slope, and $2.59$~m/s on grass. Before release, the command is smoothly reduced, allowing the controller to transition from acceleration to braking, release, and recovery without an abrupt change in the whole-body state. The mean absolute velocity-tracking errors are $0.13$, $0.34$, $0.57$, and $0.19$~m/s for the plane, stone road, slope, and grass, respectively, all remaining below $0.6$~m/s. Although the slope shows the largest deviation, this remains within an acceptable range given its lower achievable forward speed and stronger pitch variation, indicating robust command tracking under terrain disturbances. Overall, the policy preserves the intended phase structure of the run-and-throw motion across terrains, showing that locomotion and throwing are coordinated as a single phased behavior.

\section{CONCLUSIONS}

We presented a learning-based whole-body kinetic coordination framework for terrain-robust long-range throwing with a wheeled-quadruped manipulator. By coupling an SQP-optimized kinetic-chain arm reference with a history-guided whole-body RL policy, the system coordinates run-up, release, and recovery under terrain disturbances. Simulation and hardware results show that kinetic-chain optimization improves release performance, while whole-body coordination strengthens body-assisted energy transfer and preserves the intended run-and-throw phase structure across diverse terrains.



\end{document}